\documentclass[11pt]{article}

\usepackage{acl}
\usepackage{times}
\usepackage{latexsym}

\usepackage[T1]{fontenc}

\usepackage[utf8]{inputenc}

\usepackage{microtype}

\usepackage{inconsolata}

\usepackage{graphicx}

\usepackage{algorithm}
\usepackage{algorithmic}
\usepackage{multirow}
\usepackage{bbding}
\usepackage{caption}

\usepackage{booktabs}
\usepackage{siunitx}

\PassOptionsToPackage{normalem}{ulem}
\usepackage[normalem]{ulem}
\useunder{\uline}{\ul}{}
\title{ODUTQA-MDC: A Task for Open-Domain Underspecified Tabular QA with Multi-turn Dialogue-based Clarification}

\author{
  \textbf{Zhensheng Wang\textsuperscript{1,2}},
  \textbf{Zhanteng Lin\textsuperscript{3}},
  \textbf{Wenmian Yang\textsuperscript{2,\textdagger}},
  \textbf{Kun Zhou\textsuperscript{1,2}},
  \textbf{Yiquan Zhang\textsuperscript{2}},
  \textbf{Weijia Jia\textsuperscript{2,4}}
  \\
  \textsuperscript{1}School of Artificial Intelligence, Beijing Normal University, Beijing, PR China\\
  \textsuperscript{2}Institute of Artificial Intelligence and Future Networks, Beijing Normal University, Zhuhai, PR China\\
  \textsuperscript{3}Faculty of Arts and Sciences, Beijing Normal University, Zhuhai, PR China\\
  \textsuperscript{4}Beijing Normal-Hong Kong Baptist University, Zhuhai, PR China
  \\
  \small{
    \{jensenwang, zhantenglin, zhoukun\}@mail.bnu.edu.cn, 
    \{wenmianyang, jiawj\}@bnu.edu.cn, 
    zhangyq987@hotmail.com
  }
}

\newcommand{\algorithmicparameter}{\textbf{Parameter:}}
\newcommand{\PARAMETER}{\item[\algorithmicparameter]}

\begin{document}
\maketitle

\begingroup
\renewcommand\thefootnote{\textdagger}
\footnotetext{Corresponding author.}
\endgroup

\begin{abstract}

The advancement of large language models (LLMs) has enhanced tabular question answering (Tabular QA), yet they struggle with open-domain queries exhibiting underspecified or uncertain expressions. To address this, we introduce the ODUTQA-MDC task and the first comprehensive benchmark to tackle it. This benchmark includes: (1) a large-scale ODUTQA dataset with 209 tables and 25,105 QA pairs; (2) a fine-grained labeling scheme for detailed evaluation; and (3) a dynamic clarification interface that simulates user feedback for interactive assessment. We also propose MAIC-TQA, a multi-agent framework that excels at detecting ambiguities, clarifying them through dialogue, and refining answers. Experiments validate our benchmark and framework, establishing them as a key resource for advancing conversational, underspecification-aware Tabular QA research. The data and code are available at \url{https://github.com/jensenw1/ODUTQA-MDC}.

\end{abstract}

\section{Introduction}



Large language models (LLMs) have driven significant progress in open-domain tabular question answering (Tabular QA). Distinct from closed-domain settings where the target table is predetermined, the open-domain paradigm necessitates autonomously retrieving relevant tables from a large-scale database \cite{LLM4TQA:survey,Opentab:nips2024,text2SQL:ACL2023}. However, existing text-to-SQL approaches often falter when applied to this pipeline, as real-world user queries are frequently underspecified due to spelling errors, unclear expressions, or incomplete information. Such ambiguity fundamentally hinders the generation of correct SQL queries, leading to inaccurate answers.

Existing research on this problem, while valuable, often remain limited to closed-domain scenarios that focus merely on detecting and classifying underspecifications \cite{AmbiQT:EMNLP2023}. These approaches do not resolve the core challenge: generating correct answers without user clarification. Even conversational datasets like PRACTIQ \cite{PRACTIQ:NAACL2025} rely on predetermined, static dialogues, failing to capture the dynamic and unpredictable nature of real-world user interactions.

\begin{figure}[t]
 \centering
 \includegraphics[width=0.95\columnwidth]{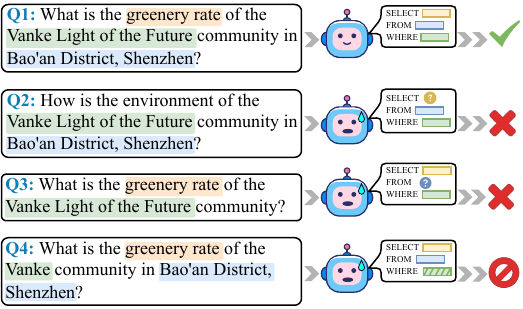}
 \centering
 \caption{Examples of underspecified input causing questions to be unanswerable.}
 \label{fig:Query_Ambigous}
\end{figure}

Real-world open-domain Tabular QA presents several core challenges that current methods do not address: autonomously selecting relevant tables from large databases, handling queries with multiple types of underspecification simultaneously, and generating accurate answers by integrating information gathered over users' multi-turn clarification dialogues \cite{DBLP:journals/chinaf/LiuWLNWC25,DBLP:journals/fcsc/QinCFWZLLCY26,DBLP:conf/acl/YeYL00025}. Progress has been limited by the absence of suitable datasets and evaluation frameworks designed for these complexities.

To address these gaps, we introduce ODUTQA-MDC, a new task for Open-Domain Underspecified Tabular Question Answering with Multi-turn Dialogue-based Clarification. The task is supported by three components: a large-scale dataset, a fine-grained labeling scheme, and a simulated clarification interface. We construct ODUTQA, the first dedicated open-domain underspecified Tabular QA dataset, with 209 structured tables and 25,105 QA pairs from property, real estate finance, and land auction domains. To align with text-to-SQL workflows, we categorize underspecifications by their correspondence with SQL structures: table-scope underspecification (impacting the FROM clause; see Figure~\ref{fig:Query_Ambigous}-Q3), query-intent underspecification (affecting the SELECT clause; see Figure~\ref{fig:Query_Ambigous}-Q2), query-condition underspecification (affecting the WHERE clause; see Figure~\ref{fig:Query_Ambigous}-Q4), and mixed underspecification.

For a nuanced evaluation, our framework introduces fine-grained underspecification detection labels. Building on intent and slot labels from spoken language understanding, following previous work \cite{RETQA:AAAI2025}, we define three auxiliary detection labels: intent underspecification (identified via binary classification), scope underspecification, and condition underspecification (both using triple-based annotations for content, category, and error type). To simulate practical interactions, we integrate a conversational agent that dynamically generates clarification based only on the underspecifications detected by the system. This establishes a dynamic “detection-clarification-redetection” closed-loop evaluation process, realistically simulating multi-turn interactions and enabling a rigorous assessment of a system's ability to handle ambiguity.

Building on this task, dataset, and evaluation framework, we introduce MAIC-TQA, an LLM-powered Multi-Agent Interactive Clarification framework for ODUTQA-MDC. MAIC-TQA comprises four collaborative agents: the Spoken Language Understanding (SLU) module, Scope Validator Agent, Table Retrieval Agent, and SQL Generation and Validation Agent. Working in concert, these agents deliver end-to-end, dynamic tabular QA with multi-turn clarification. The framework leverages external SQL tools and integrates execution feedback, such as SQL errors and empty results, to drive effective underspecification detection, adaptive clarification, and robust reasoning. Throughout the workflow, the system dynamically solicits clarifications, incorporates user responses, and iteratively refines both understanding and retrieval. By jointly optimizing underspecification identification and QA accuracy, MAIC-TQA establishes the first comprehensive agent-based benchmark for open-domain underspecified tabular QA with interactive, multi-turn dialogue.

The main contributions of this paper are as follows:
\begin{itemize}
    \item We introduce the novel task of Open-Domain Underspecified Tabular QA with Multi-turn Dialogue-based Clarification, supported by the first corresponding dataset (ODUTQA), detailed detection labels, and a dynamic conversational clarification interface.

    \item We propose an agent-based open-domain Tabular QA system capable of effectively identifying and dynamically clarifying underspecifications through intelligent multi-turn interactions, substantially enhancing system adaptability in complex scenarios.

    \item We systematically evaluate mainstream LLMs’ underspecification detection and QA performance on our dataset, presenting the first benchmark that comprehensively assesses the entire ``information detection–multi-turn clarification–QA reasoning" workflow, facilitating further development in this research area.
\end{itemize}

\section{Related Works}

\begin{table*}[ht]\small\centering
\begin{tabular}{ccccccc}
\hline
Dataset & Open Domain & Underspecification & \begin{tabular}[c]{@{}c@{}}Mixed  \\ Underspecification\end{tabular} & QA Evaluation & Conversational & \begin{tabular}[c]{@{}c@{}}Dynamic\\ Clarification\end{tabular} \\ \hline
Spider & \XSolidBrush & \XSolidBrush & \XSolidBrush & \Checkmark & \XSolidBrush & \XSolidBrush \\
RETQA & \Checkmark & \XSolidBrush & \XSolidBrush & \Checkmark & \XSolidBrush & \XSolidBrush \\
PRACTIQ & \XSolidBrush & \Checkmark & \XSolidBrush & \XSolidBrush & \Checkmark & \XSolidBrush \\
AmbiQT & \Checkmark & \Checkmark & \XSolidBrush & \Checkmark & \XSolidBrush & \XSolidBrush \\
Ambrosia & \Checkmark & \Checkmark & \XSolidBrush & \XSolidBrush & \XSolidBrush & \XSolidBrush \\
ODUTQA & \Checkmark & \Checkmark & \Checkmark & \Checkmark & \Checkmark & \Checkmark \\ \hline
\end{tabular}
\caption{Dataset comparison. ``Open Domain'' indicates table retrieval from a large corpus, whereas ``Closed Domain'' (marked with `\XSolidBrush') specifies QA on given tables.}
\label{Dataset comparision table}
\end{table*}

\subsection{TQA Datasets}

Mainstream Text-to-SQL datasets, such as Spider \cite{Spider:EMNLP2018}, WikiSQL \cite{WikiSQL:2018}, and Open-WikiTable \cite{Open-WikiTable:ACL2023}, rarely consider underspecification in user queries. RETQA \cite{RETQA:AAAI2025} provides SLU annotations but does not address underspecification. More recent datasets introduce underspecification via schema modification \cite{AmbiQT:EMNLP2023}, multi-turn dialogue \cite{PRACTIQ:NAACL2025}, or human annotation \cite{AMBROSIA:NIPS2024}. However, these datasets remain simplistic, lack systematic clarification annotation, and do not capture complex real-world mismatches between queries and database schemas. A detailed comparison of the datasets is provided in Table \ref{Dataset comparision table}.

\subsection{TQA Methods}

Existing methods address open-domain TQA and underspecification separately. SLUTQA \cite{RETQA:AAAI2025} enhances LLMs with intent-based example selection but does not resolve underspecification. PURPLE \cite{PURPLE:ICDE2024} prunes schema using a Steiner Tree approach, while LogicalBeam \cite{AmbiQT:EMNLP2023} generates diverse SQL templates, though both have limited flexibility and coverage. Most importantly, current TQA methods lack error identification and dynamic conversational clarification, limiting their effectiveness in real-world underspecified scenarios.

\section{Task Construction and Analysis }

\subsection{Task Definition}
The ODUTQA-MDC task evaluates open-domain tabular QA systems on their ability to resolve query ambiguities via multi-turn clarification. Given a user query and a multi-table database, the task consists of two stages:
\begin{itemize}
    \item \textbf{Underspecification Detection:} Identifying the specific ambiguity type within the query (i.e., SELECT, FROM, WHERE, or Mixed).
    \item \textbf{Clarification-based Dynamic QA:} Interacting with a user simulator to resolve detected ambiguities, followed by standard Table Retrieval and SQL Generation to derive the final answer.
\end{itemize}

\subsection{Datasets Construction}

\subsubsection{Data Source}
Our dataset originates from RETQA \cite{RETQA:AAAI2025}, a specified, open-domain QA dataset with 90 question templates, covering the Land Auction, Property, and Real Estate Finance domains. To support the four defined underspecification types (table-scope, query-intent, query-condition, and mixed), the original tables are normalized. Property and Land Auction tables are reorganized by city into 102 tables (avg. 1,040.37 rows) and 103 tables (avg. 30.35 rows), respectively. The four finance tables are consolidated by year (2019–2022), each averaging 444.75 rows. All resulting tables are stored in a PostgreSQL database \cite{postGRE} to simulate realistic tabular QA scenarios.

\subsubsection{Template Design and QA Pair Generation}
To construct a dataset of underspecified queries, we adapt 90 seed templates from RETQA, each associating a natural language question with a corresponding SQL query. Rather than altering the database schema, our core strategy injects underspecification by manipulating the question templates and their assigned values.

We begin with a detailed manual analysis, mapping each template variable to its SQL clause counterpart. This process connects variables to table captions (affecting the FROM clause), column headers (impacting the SELECT clause and defining user intent), and query conditions (appearing in the WHERE clause).

This mapping informs two distinct strategies for introducing underspecification. The first, template restructuring, modifies the templates themselves to induce SELECT clause underspecification and create the Missing type for the FROM clause. The second, value modification, alters values in generated QA pairs to create WHERE clause underspecification and to produce the Unmatch and Error types for the FROM clause.

These strategies yield three distinct categories of underspecification.

\begin{itemize}
    \item \textbf{SELECT Underspecification Generation:} This arises from an insufficiently specified user intent, making it difficult to identify the target column. To induce this, we replace explicit column references in templates with vague expressions. For instance, "green coverage ratio" is reformulated as ``how is the environment?'', while a question regarding ``risk level'' is modified to ``how is the business status?''. All original headers are preserved in a SELECT-Clarification dictionary for subsequent resolution.
    
    \item \textbf{FROM Underspecification Generation:} This results from missing or incorrect scope-defining information, hindering table identification. It is introduced two ways. Missing-type underspecification is generated at the template level by omitting variables such as ``city'' or ``district''. Error-type (typos) and Unmatch-type (logical conflicts, e.g., a non-existent city-district pair) are introduced by modifying values post-generation. All original, correct information is recorded in a FROM-Clarification dictionary.

    \item \textbf{WHERE Underspecification Generation:} This occurs when incomplete or erroneous conditions hinder accurate data matching. We simulate this by post-processing QA pairs, replacing lengthy proper nouns (e.g., project names) with natural abbreviations from a few-shot prompted LLM. After creating a name-to-abbreviation map, we systematically replace the proper nouns in the questions. All original full names are stored in a WHERE-Clarification dictionary for downstream use.
\end{itemize}

From the initial 90 templates, we identify 76 that are suitable for underspecification injection. These are systematically expanded into 222 extended templates, each incorporating diverse underspecification patterns in the SELECT and FROM clauses. Throughout this process, the corresponding target SQL template remains unchanged, as it represents the definitive query intent once underspecification is resolved.

To ensure data quality, we sample two QA pairs per template for rigorous manual review against criteria of fluency, consistency, underspecification preservation, and annotation correctness. Templates are iteratively revised until they meet these standards.


In the generation stage, validated templates are populated with sampled database entities, yielding 63,496 initial QA pairs. We then perform stratified sampling using the 76 template groups as strata, since templates naturally define distinct query structures and underspecification patterns and therefore provide an appropriate unit for controlling data balance. During this process, we further apply quality-control filters before and after SQL execution to remove cases that are unsuitable for an answerable QA benchmark. We then sample according to per-group target counts to achieve a balanced distribution across well-specified, single-underspecified, and mixed-underspecified examples. This procedure produces a final dataset of 25,105 QA pairs, including 5,095 well-specified, 7,892 single-underspecified (`SELECT': 197; `FROM': 3,274; `WHERE': 4,421), and 12,118 mixed-underspecified instances. The data is split 6:2:2 into training (14,973), validation (5,046), and test (5,086) sets (see Appendix \ref{appendix:Dataset Statistics}).

To further enhance linguistic naturalness, we paraphrase questions with an LLM. Given our dataset’s focus on underspecification, we use prompt engineering to explicitly prohibit the LLM from using forbidden words, namely, column headers specified in the SELECT-Clarification dictionary, thus preventing underspecified expressions from being made explicit. See Appendix \ref{appendix:LLM-based Query Rewriting} for details.

\subsection{Label Annotation}

To support fine-grained analysis and evaluation of underspecification, we design a unified annotation scheme comprising two main categories: labels for spoken language understanding (SLU) \cite{xing2025dxa,DBLP:conf/aaai/0001CZW0CL25} and labels for underspecification detection. Specifically, following previous work \cite{RETQA:AAAI2025}, we employ a template-filling approach to synthesize the dataset, where all corresponding annotations are automatically inferred during the generation process. More details of label generation are provided in Appendix \ref{sec:Dataset Construction and Statistics}.

\subsubsection{Intent and Slot Annotation}

For SLU, we annotate both intent and slot information for each question. Specifically, following previous work~\cite{RETQA:AAAI2025}, each seed template is assigned one or more intent labels. In cases where underspecification is introduced through the SELECT clause, the intent is annotated as ``Unknown'' to reflect uncertainty in user intent.

Slot annotation adopts the standard BIO (Begin, Inside, Outside) format. Variables within question templates are treated as slots, with their types defined by the corresponding variable names, while all other tokens are labeled as ``O''. 

\subsubsection{Underspecification Type Annotation}
In addition to standard annotations, our dataset defines three auxiliary underspecification detection labels: intent underspecification, scope underspecification, and condition underspecification, as follows:
 
\begin{itemize}
\item \textbf{Intent Underspecification Label:} This binary label indicates whether a query exhibits intent underspecification. It is assigned as ``True'' if the user intent is unclear (i.e., with ``unknown'' intent label ), and ``False'' if the user intent is clear (i.e., with specific intent labels).

\item \textbf{Scope Underspecification Label:} This label employs a triple-based annotation format [``slot\_content'', ``slot\_type'', ``error\_type''], capturing errors related to table scope. For example, if a query omits the required ``City'' slot, the label is [``'', ``City'', ``Missing'']. The possible values for ``error\_type'' are as follows:
\begin{itemize}
    \item \textit{Missing}: A required slot is absent from the query.
    \item \textit{Error}: The slot content cannot be found in the database.
    \item \textit{Unmatch}: The individual slot values are valid, but their combination does not exist in the database.
\end{itemize}
These labels directly determine whether the correct database table can be identified and serve as explicit signals for triggering system clarification.

\item \textbf{Condition Underspecification Label:} This label addresses underspecifications in the WHERE clause, particularly cases where a provided entity name (such as a project or enterprise) does not exist in the database. In such instances, the system generates a triplet [``slot\_content'', ``slot\_type'', ``not exist''], for example, [``ABC Technology Inc.'', ``enterprise name'', ``not exist'']. This enables the system to explicitly identify invalid query conditions and initiate targeted clarification or correction.
\end{itemize}

\subsection{Conversational Clarification Interface}
\label{Conversational Clarification Interface}

To overcome the limitations of prior work that focuses on static underspecification classification, we introduce an automated interactive user simulator that resolves underspecified queries through multi-turn clarification dialogues. This design choice addresses a critical practical constraint: while human interaction provides high realism, it is prohibitively expensive and lacks consistency and reproducibility at scale. Our simulator offers a scalable and standardized alternative that preserves sufficient linguistic realism for robust evaluation.


The simulator is implemented as a callable Python interface for handling SELECT, FROM, and WHERE underspecification types. Clarification is strictly gated by detection accuracy: corrective information is provided only when the predicted underspecification label matches the ground truth; otherwise, no clarification is revealed. Upon correct detection, the interface retrieves the corresponding ground-truth value from clarification dictionaries. This value is inserted into a standardized response template.

To emulate natural user interactions, an LLM is employed to paraphrase these templates into colloquial expressions. To guarantee reliability, we implement a rigorous post-generation verification mechanism. Specifically, the system validates that the ground-truth value retrieved from the dictionary remains invariant within the paraphrased response. If the value is missing, the system regenerates the response; if regeneration fails after a fixed number of attempts, it falls back to the standardized template. This design ensures linguistic diversity without sacrificing factual accuracy.


To support diverse benchmarking, the interface offers two modes: Dynamic Mode generates varied responses to assess robustness under realistic conditions, while Fixed Mode delivers standardized responses for reproducibility. 
See Appendix \ref{sec:interface_details} for implementation details and Appendix \ref{sec:appendix_clarification_rationale} for the design rationale.

\begin{figure*}[hbpt]
 \centering
 \includegraphics[width=0.95\textwidth]{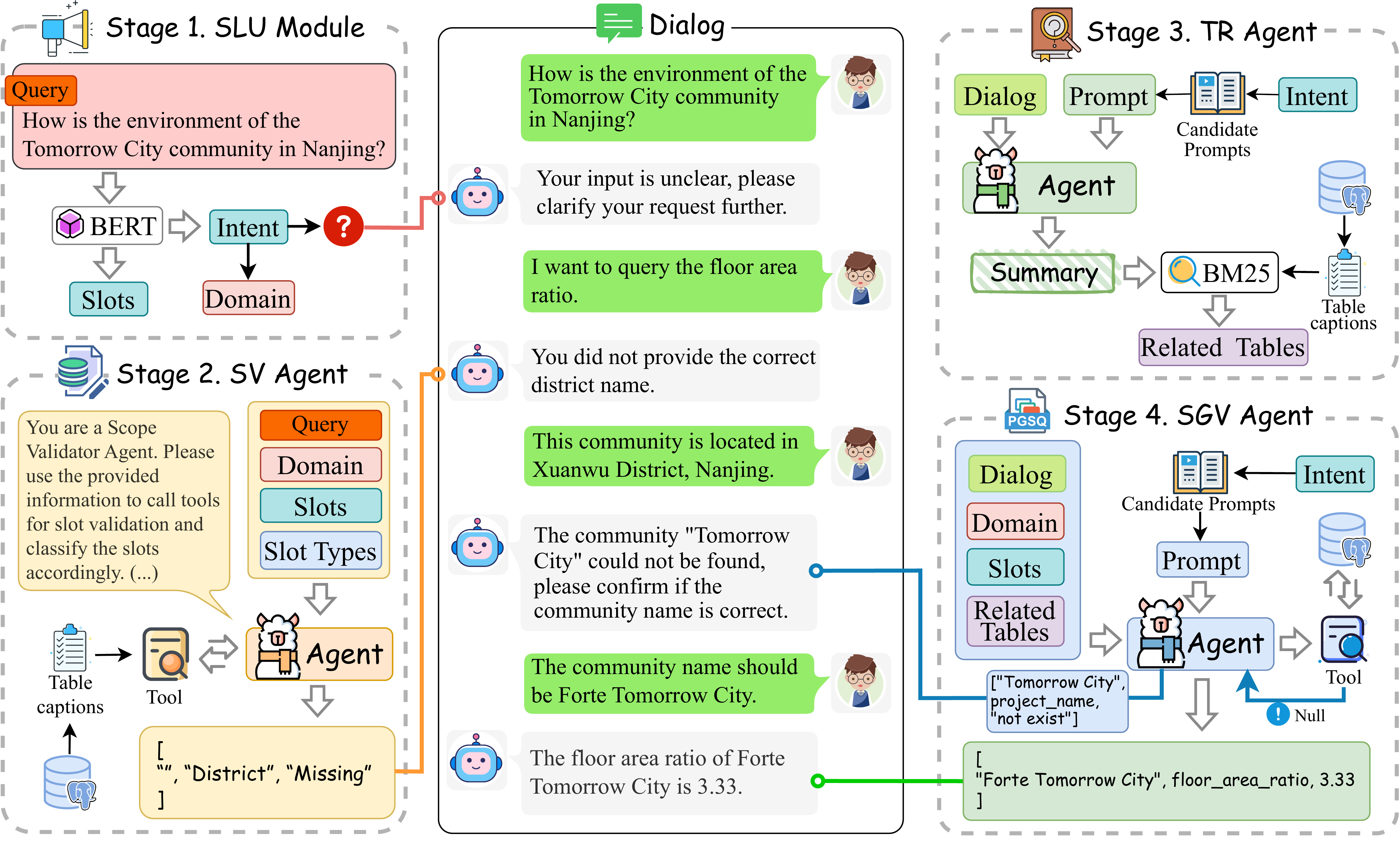}
 \centering
 \caption{General framework of MAIC-TQA.}
 \label{fig:Workflow}
\end{figure*}

\section{Method}

\subsection{Overview}

As illustrated in Figure~\ref{fig:Workflow}, the MAIC-TQA framework adopts a modular, multi-agent architecture to address open-domain underspecified tabular question answering with multi-turn clarification. The workflow proceeds as follows: First, the SLU Module (Section \ref{SLUM}) extracts user intent and slot information from the natural language query. Next, the Scope Validator Agent (Section \ref{SV-Agent}) verifies and, if necessary, requests clarification for caption-related slots to ensure accurate table scope identification. The Table Retrieval Agent (Section \ref{TR-Agent}) then integrates the original query and user clarifications to determine the most relevant table caption. Finally, the SQL Generation and Validation Agent (Section \ref{SGV-Agent}) constructs and executes the SQL query, validating the output to generate the final answer \cite{duan-etal-2025-orpp}.

\subsection{SLU Module}
\label{SLUM}
Given a user’s natural language query, we first employ a BERT-based classifier \cite{BERT:NAACL2019} for intent detection and slot filling \cite{qin2021co,cheng2023scope}. An intent underspecification label is generated based on the classifier’s output. If intent underspecification is detected, the system issues a template-based follow-up question (e.g., ``Your input is unclear. Please clarify your request.'') to prompt user clarification. Upon receiving the user’s response, the system concatenates the original query with the clarification to form an updated conversational context, which is then reprocessed by the BERT classifier for a second attempt at intent recognition. Once a valid intent is identified, the system proceeds to slot validation. To maintain interaction efficiency, the clarification loop is restricted to a single iteration.

For downstream processing, clarified intent labels are mapped to one of three domains, i.e., Property Sales, Land Auctions, or Enterprise Finance, as the domain labels for subsequent tasks.

\subsection{Scope Validator Agent}\label{SV-Agent}
The primary function of the Scope Validator Agent (SV Agent) is to verify the scope information in user queries, enabling accurate table matching. Scope information is typically represented by slot annotations; therefore, the agent receives the user’s original query, the domain label, the slot labels, and the list of domain-required slots as input.

The validation process begins by checking for missing slots. The agent compares the extracted slots against the required slot types for the given domain. If a required slot is missing, the agent immediately assigns a scope underspecification label with ``Missing'' as the \textit{error\_type}, following the specified prompt. For example, if the ``District'' slot is absent, the output triplet would be (``'', ``District'', ``Missing''), as shown in Figure \ref{fig:Workflow}.

If all required slots are present, the agent proceeds to verify the validity of the provided slot values. This is accomplished by calling a validation function, which cross-references the input slot values with the corresponding database tables for the specified domain, based on a detailed prompt. If any slot value fails validation (e.g., the content does not exist or the slot combination is invalid), the agent generates the appropriate Scope Underspecification Label and prompts the user for clarification.

This two-stage validation process ensures both the completeness and the correctness of scope information, facilitating robust and accurate table selection. The detailed prompt templates and validation functions used by this agent are provided in Appendix \ref{appendix:Prompts and Functions of Agent}.

\subsection{Table Retrieval Agent}\label{TR-Agent}
The Table Retrieval Agent (TR Agent) generates a concise summary of the target table based on the user’s original query and the supplementary information collected by the Scope Validator Agent. It analyzes the complete dialogue history, identifying system prompts for clarification and corresponding user responses. According to the previously predicted intent label, the agent selects different representative examples as context for in-context learning. It then integrates information from both the initial query and the clarifications to produce a table summary, which serves as the predicted table caption.

Table retrieval follows a two-stage process. First, the system attempts to find an exact match between the generated summary and the table captions in the database. If there is no exact match, the system performs a similarity search using the BM25 algorithm \cite{BM25:2009}. If the agent cannot generate a summary due to insufficient dialogue information, the full dialogue history is used directly for BM25 retrieval. This approach ensures robust table retrieval in all cases. The detailed prompt templates used by this agent are provided in Appendix \ref{appendix:Prompts and Functions of Agent}.

\subsection{SQL Generation and Validation Agent}\label{SGV-Agent}
After table retrieval, the SQL Generation and Validation Agent (SGV Agent) is responsible for generating and validating SQL queries. The agent takes four inputs: the complete dialogue history, the predicted domain, the extracted slots, and the related table captions. Guided by the user's intent, the agent selects an appropriate prompt template containing five representative gold SQL examples for in-context learning, and then generates and executes the SQL query accordingly.

If the query is successful and meets all user requirements, the system returns the final SQL query and result, completing the task. If the result is empty or incomplete (e.g., only a subset of the requested entities is returned), the agent calls a predefined SQL validation function to identify missing conditions, generates a condition underspecification label, and requests clarification from the user. This clarification is appended to the dialogue history to form an updated context for a single round of SQL regeneration. If the new query still fails or returns no results, the process ends, and the empty output is recorded as the final result.

Details of the prompt templates and the SQL validation function are provided in Appendix \ref{appendix:Prompts and Functions of Agent}.

\section{Experiments}
\label{sec:Experiments}

\subsection{Tasks and Baselines}
We utilize ODUTQA-MDC as the benchmark for open-domain underspecified tabular question answering, which consists of two sub-tasks: (i) underspecification detection, including intent, scope, and condition detection, and (ii) clarification-based dynamic QA that evaluates a system’s ability to resolve underspecified queries through interaction. As this benchmark explicitly targets clarification-aware reasoning over incomplete user inputs, there are no directly comparable prior systems that jointly model underspecification detection and clarification-driven query execution in tabular settings.

For comparative evaluation, we implement MAIC-TQA together with a state-of-the-art tabular QA framework, SLUTQA~\cite{RETQA:AAAI2025}, which does not employ clarification and instead answers directly from underspecified queries, serving as a clarification-free baseline. For MAIC-TQA, we evaluate both fixed and dynamic clarification responses, where all dynamic responses are generated by Qwen2.5-72B for consistency. The SGV Agent is responsible for SQL generation and execution, taking the complete dialogue history, predicted domain, extracted slots, and table captions as input; guided by user intent, it selects a prompt template with five gold SQL examples for in-context learning and generates the executable query. We support both MAIC-TQA and SLUTQA using LLM backbones with function-calling capabilities, including open-source Qwen3 32B and Qwen3 30B (A3B)\cite{Qwen3:report} deployed via SGLang\cite{SGlang:nips2024} with default chain-of-thought, and closed-source Kimi K2~\cite{kimiteam2025}, GLM 4~\cite{glm2024chatglm}, and Doubao (Seed 1.6 flash-250828) accessed through official APIs with default settings. Additional details are provided in Appendix \ref{appendix:Implementation of baselin} and \ref{appendix:Computing Infrastructure Statement}.

\begin{table*}[t]\centering
\footnotesize
\begin{tabular}{c | cc cc cc c | cc cc c}
\hline
 & \multicolumn{7}{c|}{\textbf{Fixed Clarification}} & \multicolumn{5}{c}{\textbf{Dynamic Clarification}} \\
Model
 & \multicolumn{2}{c}{SELECT}
 & \multicolumn{2}{c}{FROM}
 & \multicolumn{2}{c}{WHERE}
 & Mixed
 & \multicolumn{2}{c}{FROM}
 & \multicolumn{2}{c}{WHERE}
 & Mixed \\
 & Acc. & F1
 & Acc. & F1
 & Acc. & F1
 & Acc.
 & Acc. & F1
 & Acc. & F1
 & Acc. \\ \hline
BERT
 & 99.78 & 99.22
 & -- & --
 & -- & --
 & --
 & -- & --
 & -- & --
 & -- \\ \hline
Qwen3 32B
 & -- & --
 & 77.66 & 82.82
 & 69.59 & 66.02
 & 54.96
 & 78.78 & 85.11
 & {\ul 73.81} & {\ul 72.49}
 & {\ul 59.42} \\ \hline
Qwen3 30B
 & -- & --
 & 75.17 & 85.10
 & \textbf{75.67} & \textbf{78.99}
 & {\ul 58.55}
 & 69.95 & 77.45
 & 72.49 & 69.50
 & 52.92 \\ \hline
Kimi K2
 & -- & --
 & \textbf{82.60} & {\ul 87.95}
 & 69.02 & 65.54
 & 55.51
 & \textbf{81.63} & {\ul 87.31}
 & 64.81 & 59.55
 & 52.69 \\ \hline
GLM 4
 & -- & --
 & 81.10 & 86.71
 & 44.77 & 28.96
 & 37.97
 & 78.55 & 84.39
 & 57.02 & 53.41
 & 46.23 \\ \hline
Doubao
 & -- & --
 & {\ul 81.36} & \textbf{89.46}
 & {\ul 73.72} & {\ul 75.69}
 & \textbf{61.21}
 & {\ul 81.22} & \textbf{89.23}
 & \textbf{83.59} & \textbf{84.38}
 & \textbf{68.50} \\ \hline
\end{tabular}
\caption{Overall performance of underspecification detection under fixed and dynamic clarification settings.}
\label{tab:underspec_overall}
\end{table*}

\begin{table}[t]\centering\footnotesize
\begin{tabular}{c|c|c|cc}
\hline
\multirow{2}{*}{Model} & \multirow{2}{*}{Method} & \begin{tabular}[c]{@{}c@{}}\textbf{Table}\\ \textbf{Retrieval}\end{tabular} & \multicolumn{2}{c}{\begin{tabular}[c]{@{}c@{}}\textbf{SQL}\\ \textbf{Generation}\end{tabular}} \\ \cline{3-5} 
 &  & F1 & ECR & EA \\ \hline
\multirow{3}{*}{Qwen3 32B} & SLUTQA & 60.07 & 85.90 & 24.20 \\
 & Ours (Fix) & 92.89 & 97.46 & 57.44 \\
 & Ours (Dyn) & \textbf{93.48} & \textbf{98.13} & \textbf{57.89} \\ \hline
\multirow{3}{*}{Qwen3 30B} & SLUTQA & 58.78 & 86.75 & 21.22 \\
 & Ours (Fix) & \textbf{95.10} & 96.32 & \textbf{54.11} \\
 & Ours (Dyn) & 91.17 & \textbf{97.56} & 49.55 \\ \hline
\multirow{3}{*}{Kimi-K2} & SLUTQA & 62.33 & 88.35 & 30.70 \\
 & Ours (Fix) & 95.07 & 98.26 & \textbf{63.89} \\
 & Ours (Dyn) & \textbf{97.13} & \textbf{98.39} & 59.86 \\ \hline
\multirow{3}{*}{GLM 4} & SLUTQA & 60.15 & 85.51 & 25.75 \\
 & Ours (Fix) & \textbf{89.89} & 85.88 & 32.47 \\
 & Ours (Dyn) & 89.28 & \textbf{88.84} & \textbf{44.63} \\ \hline
\multirow{3}{*}{Doubao} & SLUTQA & 59.95 & 78.52 & 23.72 \\
 & Ours (Fix) & 77.40 & 90.81 & 41.54 \\
 & Ours (Dyn) & \textbf{93.43} & \textbf{97.56} & \textbf{53.23} \\ \hline
\end{tabular}
\caption{Table retrieval and SQL generation results (``Fix''/``Dyn'': Fixed/Dynamic clarification).}
\label{Table retrival and SQL Result}
\end{table}

\subsection{Implementation}





This section presents the evaluation metrics and implementation details for the two core tasks defined in Section 3.4: Underspecification Detection and Clarification-based Dynamic QA.

\noindent\textbf{Evaluation Metrics}
For the Underspecification Detection task, we adopt Accuracy and macro F1-score to assess the model's ability to detect intent, scope, and condition underspecifications. Accuracy is calculated on the entire dataset, while F1 is reported specifically for underspecified queries.

For the Clarification-based Dynamic QA task, evaluation is divided into two sub-tasks: (1) Table Retrieval, evaluated using F1 score; and (2) SQL Generation, assessed by Execution Correctness Rate (ECR) and Execution Accuracy (EA) \cite{Spider:EMNLP2018}. ECR denotes the proportion of generated SQL statements that can be executed without error, while EA measures the proportion of cases where the execution result of the generated SQL exactly matches the gold standard result.

For SLUTQA, BERT parameters are loaded directly from the officially released pre-trained model. For our method, BERT is fine-tuned on the ODUTQA training set. All LLM API calls follow their respective official default settings. Detailed performance of the SLU Module is provided in Appendix \ref{appendix:SLU Task}.

\subsection{Results and Discussions}
\label{subsec:Results and Discussions}
In this section, we present the experimental results and analysis. The results of the ablation study are shown in Appendix \ref{appendix:Ablation Study}. 

For the underspecification detection task, we evaluate model performance on three types corresponding to SQL clauses: intent (SELECT), scope (FROM), and condition (WHERE). As summarized in Table \ref{tab:underspec_overall}, the BERT-based approach achieves strong results in both Accuracy and F1 score for intent underspecification. For scope underspecification, Kimi-K2 achieves the highest accuracy in both dynamic and fixed scenarios, while Doubao Seed 1.6 achieves the highest F1 score in both scenarios. For condition underspecification, Qwen3 30B A3B achieves the highest accuracy and F1 score in the fixed scenario, whereas Doubao Seed 1.6 achieves the highest accuracy and F1 score in the dynamic scenario. The mixed underspecification task proves extremely challenging due to an error accumulation effect, where failing to detect a single problem type leads to an overall incorrect prediction. Despite this difficulty, Doubao Seed 1.6 achieves the highest accuracy for this task across both dynamic and fixed scenarios.

In table retrieval, our method demonstrates robust performance across all models, as reported in Table~\ref{Table retrival and SQL Result}. This gain is primarily attributed to our method’s ability to proactively identify missing key information, such as table captions, and to elicit necessary details via follow-up questions, resulting in more precise retrieval.

For SQL generation, our method demonstrates superior performance on all models, as shown in Table \ref{Table retrival and SQL Result}. 
Specifically, on Doubao Seed 1.6, our method achieves execution compliance rate (ECR) improvements of 12.29\% and 19.04\% over SLUTQA in the fixed and dynamic scenarios, respectively. Furthermore, on the EA metric, which more accurately captures intent alignment, our method achieves substantial improvements of 33.24\% and 33.69\% on Qwen3 32B for the fixed and dynamic scenarios, respectively, which further underscores the necessity of clarification mechanisms for handling underspecified problems in table QA.

\section{Conclusion}
In this paper, we introduce ODUTQA-MDC, a new task for open-domain underspecified tabular question answering with multi-turn clarification. We provide three resources to support evaluation: ODUTQA, the first large-scale underspecified tabular QA dataset; a fine-grained underspecification labeling scheme; and a dynamic clarification interface. We further propose MAIC-TQA, an LLM-based agent framework that combines adaptive underspecification detection with iterative clarification. Extensive experiments benchmark diverse LLMs and baselines, delivering the first unified evaluation of underspecification detection, clarification, and end-to-end QA, and establishing foundational benchmarks for underspecification-aware tabular QA.


\section*{Limitations}
While our proposed framework offers significant advancements in handling underspecified user queries, we acknowledge specific boundaries in our experimental design regarding the scope of ambiguity, interaction modeling, and domain coverage. Regarding the dataset and SQL complexity, ODUTQA-MDC focuses strictly on informational underspecification within the SELECT, FROM, and WHERE clauses, intentionally excluding complex aggregations (e.g., GROUP BY or HAVING) and open-ended linguistic ambiguities. This scoping is a deliberate design choice to isolate and rigorously evaluate the model’s ability to perform logical clarification mapping without the confounding factors of linguistic noise or advanced compositional reasoning, which would obscure the core clarification mechanism in a controlled study.

Concerning the evaluation paradigm, our user simulator prioritizes logical consistency and factuality over the full spectrum of human behavioral stochasticity, such as irrelevant chitchat or dynamic goal deviations. While this reduces the chaotic nature of real-world interactions, it is critical for establishing a reproducible benchmark. By minimizing interaction noise, we ensure that performance differences can be attributed solely to the model's capacity to resolve logical ambiguities rather than its robustness to open-domain distractions, thus serving as a standardized framework for this specific capability.

Finally, with respect to domain generalizability, our current experiments are centered on the real estate market, a vertical characterized by dense attributes and user constraints. However, the proposed framework is designed to be schema-agnostic, with clarification logic grounded in universal SQL syntax rather than domain-specific rules. Therefore, despite the single-domain dataset, the underlying methodology for detecting and resolving underspecification remains methodologically generalizable to other vertical industries such as healthcare or e-commerce.

\section*{Acknowledgments}

This work is supported in part by the National Natural Science Foundation
of China (NSFC) under Grant 62272050 and the grant of Beijing Normal-
Hong Kong Baptist University sponsored
by Guangdong Provincial Department of Education;  in part by Zhuhai Science-Tech Innovation Bureau under
Grant No. 2320004002772 and the Interdisciplinary Intelligence
Super Computer Center of Beijing Normal University (Zhuhai).

\bibliography{custom}

\appendix

\section{Dataset Construction and Statistics}
\label{sec:Dataset Construction and Statistics}

\subsection{Template Filling}

\subsubsection{Implementation Details}
\begin{figure}[t]
 \centering
 \includegraphics[width=\columnwidth]{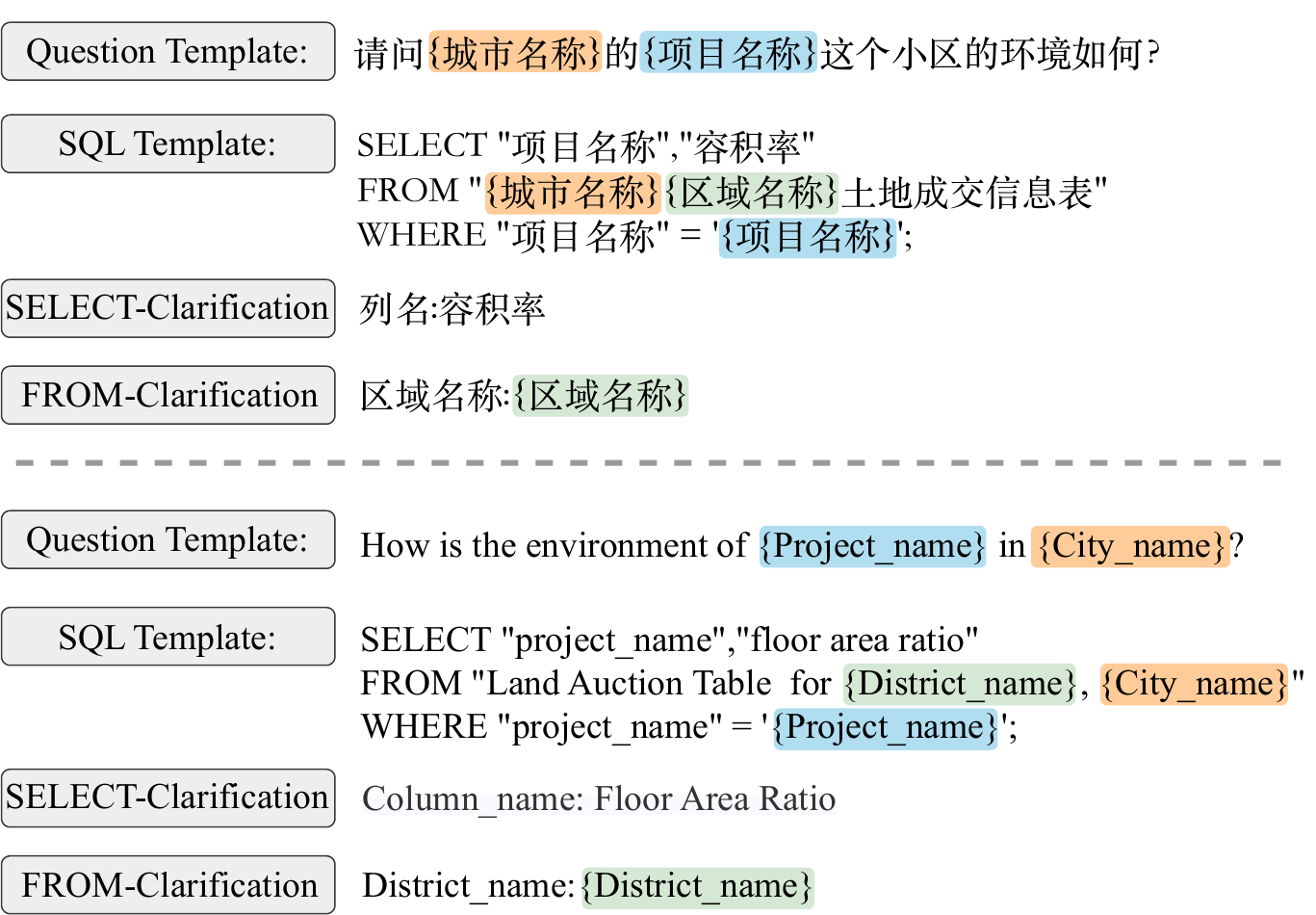}
 \centering
 \caption{Examples of Template Filling (Chinese–English Bilingual).}
 \label{fig:Template_Filling}
\end{figure}

\begin{figure}[ht]
 \centering
 \includegraphics[width=\columnwidth]{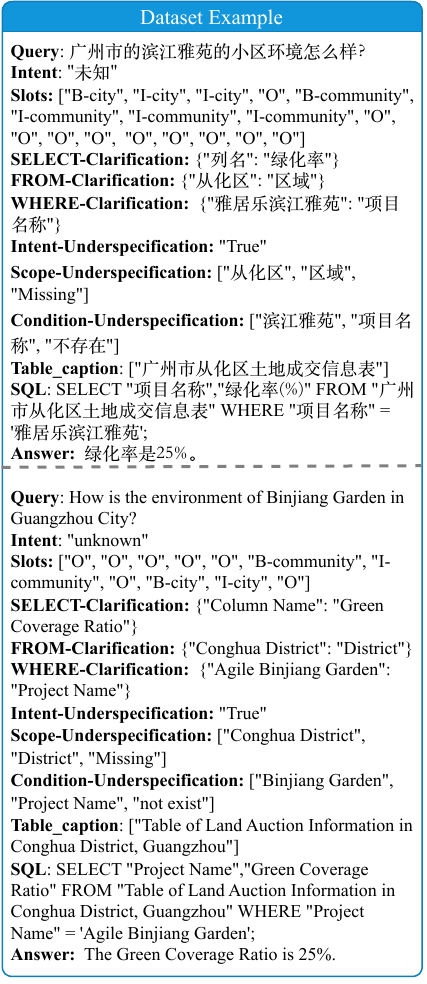}
 \centering
 \caption{QA example.}
 \label{fig:QA example}
\end{figure}

Based on 209 tables, we design 222 seed templates. As shown in Figure~\ref{fig:Template_Filling}, these templates are categorized by underspecification type: 38 with SELECT underspecification, 113 with FROM underspecification, and 71 that are specified. Each seed template comprises four components: a Question Template, an SQL Template, a SELECT clarification, and a FROM clarification. The parts enclosed in ``\{\}" are slots to be populated, and the slots across these four components are correspondingly linked. After randomly selecting appropriate values from the database, these values are used to populate the corresponding slots within the template.
In the template's components, the SELECT clarification is stored in a dictionary format without any slots; its purpose is to record the definitive user's final intent after an underspecified query is clarified. The FROM clarification contains slots related to the table caption, which are populated concurrently with other slots during the data sampling stage. Furthermore, the static text (i.e., the non-slot part) of a template determines its intent type and is uniformly labeled ``O" during BIO tagging, whereas the slot parts are tagged according to their specific sampled values.
Once all slots in a template are populated, the system executes the query generated from the SQL template, and the query result serves as the final answer to the corresponding question. An example of a complete generated QA pair is illustrated in Figure~\ref{fig:QA example}.

\subsubsection{Rationale for Template-based Construction}
Template-based construction with controlled perturbations is standard practice in recent QA and TQA datasets, such as MULTITQ \cite{DBLP:conf/acl/ChenL023}, RETQA \cite{RETQA:AAAI2025}, and SpatialRGPT-Bench \cite{DBLP:conf/nips/ChengYFGYK0L24}, etc. This approach ensures that the generated questions are logically consistent and answerable. Compared to manual annotation or crowdsourcing, which is prohibitively expensive and often suffers from inconsistent quality, template-based generation allows for high-volume, high-quality data creation that covers specific phenomena systematically.
We acknowledge that pure templates could in principle reduce diversity; therefore, we deliberately decouple semantic patterns from surface forms. Specifically, templates are used to encode the SQL intent and the specific type of underspecification. The natural-language wording is then produced by a LLM that is prompted to use varied, colloquial question styles while strictly preserving the intent and ambiguity constraints. Dictionaries are utilized only to sample realistic entities and attribute values from real tables (e.g., city names or financial indicators), not to constrain how questions are phrased. Combined with random sampling over templates, entities, and operators, this strategy yields a broad range of question types. The relatively low Text-to-SQL EA on MAIC-TQA (see Section~\ref{subsec:Results and Discussions}) demonstrates that these questions, despite being synthesized, still present significant challenges to current state-of-the-art models.

\begin{figure*}[ht]
 \centering
 \includegraphics[width=\textwidth]{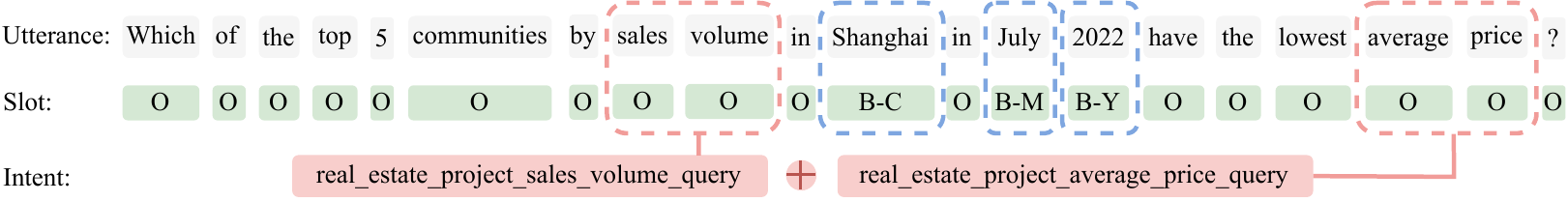}
 \centering
 \caption{Illustration of an utterance with SLU tags. The example features two intents, where `B-C' denotes `B-City' and `B-Y' denotes `B-Year'.}
 \label{fig:MDC_SLU}
\end{figure*}

\subsection{SLU Annotations in TQA}
\label{sec:slu_annotations}

Figure \ref{fig:MDC_SLU} illustrates the application of Spoken Language Understanding (SLU) tags, which serve as a critical semantic bridge between unstructured natural language and structured SQL queries in our framework. The SLU annotations consist of two distinct components: Slots and Intents.

\begin{enumerate}
    \item \textbf{Slot Tags (BIO Annotation):} We employ the BIO (Begin-Inside-Outside) tagging scheme to identify specific entities that act as filtering constraints in the SQL query. In the visual example, the tokens ``Shanghai'', ``July'', and ``2022'' are explicitly tagged as \texttt{B-C} (City), \texttt{B-M} (Month), and \texttt{B-Y} (Year), respectively. Among them, ``Shanghai'' serves as the table scope information (affecting the \texttt{FROM} clause), while ``July'' and ``2022'' map directly to the condition values in the SQL \texttt{WHERE} clause (e.g., \texttt{WHERE Date = `2022-07'}).
    
    \item \textbf{Intent Labels:} Intents capture the semantic goal of the query, determining which columns and operations (such as sorting or aggregation) are required. The example demonstrates a \textit{composite intent} scenario where the user asks for both volume ranking and price comparison. Consequently, the utterance is labeled with two intents.
\end{enumerate}

\paragraph{Role in TQA.}
In the Table Question Answering task, these SLU tags serve as critical intermediate signals for bridging natural language and structured queries. The Intent labels help the model identify the user's query goal, enabling more effective in-context example retrieval for few-shot learning. The Slot tags, on the other hand, capture the key entities that constitute essential components of the SQL query, such as filtering conditions and target columns. By explicitly decoupling entity extraction (Slots) from logic determination (Intents), the model can better understand user intent and generate SQL queries that are both syntactically correct and semantically aligned with the user's request.

\subsection{LLM-based Query Rewriting}
\label{appendix:LLM-based Query Rewriting}

Template-generated queries often display rigid and monolithic syntactic patterns, which differ markedly from the more flexible and natural language used by real users. This mismatch can lead to suboptimal model performance when handling authentic user queries. To mitigate this issue, we leverage Large Language Models (LLMs)—specifically, the Qwen2.5-72B model—to rewrite template-generated queries. This rewriting process is designed to preserve the original semantics while introducing greater linguistic diversity through synonym replacement and syntactic structure adjustments, resulting in queries that better resemble human language. Building on this, we also adopt an augmentation strategy inspired by RETQA, in which questions are further paraphrased using an LLM to enhance linguistic naturalness. Given that our dataset targets underspecified scenarios, we take care to prevent the LLM from simplifying underspecified expressions into non-underspecified forms. To this end, we enforce additional constraints in the prompt, prohibiting the use of specific forbidden words—namely, column names defined in the SELECT-Clarification dictionary—to maintain the intended underspecification. Details of the prompt formulations used for these rewriting processes are provided in Appendix \ref{appendix:Prompts and Functions of Agent}, Figure \ref{fig:rewrite prompt}.

\subsection{Quality Assessment of LLM-based Paraphrasing}
\label{sec:quality_assessment}

To rigorously validate the quality of the LLM-based rewriting, we conducted a manual assessment on samples drawn exclusively from the test set. We excluded QA pairs without underspecification and randomly selected 200 QA pairs that contained ambiguities for detailed review. The sample distribution included 30 cases of underspecification in the SELECT clause, 137 in the FROM clause, and 159 in the WHERE clause. Note that the sum exceeds 200 due to the presence of mixed-underspecification instances, which were counted across the relevant categories.

Human annotators reviewed the 200 sampled QA pairs and confirmed that the rewriting process did not compromise the integrity of the data; specifically, no instances were found where the three types of ambiguities were inadvertently deleted or resolved by the LLM. This robust preservation of ambiguity is attributed to our dual-layered validation strategy. First, our prompt design explicitly incorporates specific ``forbidden keywords''—corresponding to the missing information—to constrain the generation process and prevent the model from filling in the gaps. Second, we implemented a rigorous string matching check to verify that these prohibited terms do not appear in the final rewritten queries. This combination of constrained prompting and post-generation verification ensures the accuracy and reliability of the paraphrasing process.

\subsection{Details of Conversational Clarification Interface}
\label{sec:interface_details}

To clarify the mechanism introduced in Section \ref{Conversational Clarification Interface}, we describe the implementation details of our user simulator, which runs in two modes: Fixed and Dynamic.

1. Data Retrieval and Template Construction.
When the system correctly detects an underspecification (e.g., a missing city in the scope), the interface looks up the ground-truth value from the dataset's clarification dictionary (e.g., FROM\_clarification). It then constructs a standard response sentence, such as ``The correct city is [Beijing].''

\begin{figure*}[htbp]
 \centering
 \includegraphics[width=\textwidth]{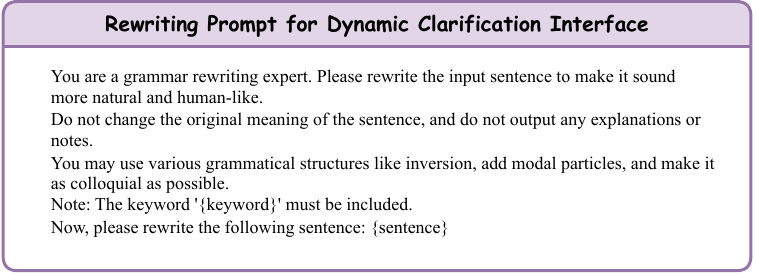}
 \centering
 \caption{Prompt for sentence rewriting used in the dynamic clarification interface.}
 \label{fig:Dynamic_Interface_Prompt}
\end{figure*}

2. Dynamic Response Generation (LLM-based Rewriting).
In the dynamic scenario, to simulate the diverse and unpredictable nature of real human language, we employ a LLM (Qwen2.5-72B) to rewrite the standard response. We utilize a specialized prompt (see Figure \ref{fig:Dynamic_Interface_Prompt}) that instructs the LLM to act as a grammar expert, making the sentence more colloquial and using varied sentence structures (e.g., inversions).

3. Reliability Control Mechanism.
A critical challenge in automatic generation is ensuring factual consistency. We address this by strict keyword extraction and validation:
\begin{itemize}
\item \textbf{Keyword Extraction}: Before generation, the system identifies the critical information slot (e.g., "Beijing") that must be present in the response.
\item \textbf{Iterative Validation}: The generation process includes a max\_retries mechanism (set to 5). After the LLM generates a rewrite, the interface automatically checks if the extracted keyword exists in the new sentence.
\item \textbf{Fallback Strategy}: If the generated sentence does not contain the keyword after 5 attempts, the system discards the LLM output and defaults to the standard template sentence.
\end{itemize}

This mechanism ensures that the ``Dynamic'' mode tests the target model's ability to understand natural language variations without introducing false information that could invalidate the QA evaluation.

\subsection{Rationale for the Simulated Clarification Interface}
\label{sec:appendix_clarification_rationale}

While we acknowledge that simulated interactions cannot fully capture the stochastic nature of human behavior, our design choice to utilize a Dynamic Clarification Interface rather than human-in-the-loop evaluation is deliberately driven by the specific requirements of constructing a robust benchmark. The rationale behind this decision is threefold:

\paragraph{Reproducibility and Standardization for Benchmarking.}
The primary objective of ODUTQA is to provide a standardized testbed for evaluating SQL generation capabilities under uncertainty. Human-based evaluation, while offering high realism, inherently suffers from high variance, subjectivity, and irreproducibility. A strictly ``real'' environment would make it mathematically impossible to fairly compare different models or track progress over time, as the ``test set'' (the variable human responses) would shift with every interaction. Our simulated approach guarantees that every model is evaluated against a consistent, reliable, and rigorous standard, which is paramount for a scientific benchmark.

\paragraph{The Hybrid Design: Controlled Logic with Generative Flexibility.}
We clarify that our interface is not ``rule-based'' in the traditional sense of rigid templates. We employ a \textbf{hybrid architecture} to balance logical correctness with linguistic naturalness:
\begin{itemize}
    \item \textbf{Decision Layer (Deterministic):} The decision of \textit{what} to clarify is governed by logic derived from the dataset's Gold Labels. This is necessary to ensure the provided information accurately resolves the specific ambiguity in the SQL generation task without introducing hallucinatory noise or irrelevant deviations.
    \item \textbf{Generation Layer (Probabilistic):} The actual dialogue utterances are generated by a state-of-the-art LLM (Qwen2.5-72B) conditioned on the dialogue history. As shown in our manual inspection, this allows for diverse wording, varying sentence structures, and natural phrasing. The model is free to formulate the question as long as it conveys the necessary disambiguating information.
\end{itemize}
This design allows us to ``stress test'' models with linguistically diverse inputs while maintaining the logical precision required for automated evaluation.

\paragraph{Scalability and Cost-Effectiveness.}
For a benchmark to be practically useful to the research community, it must be scalable. Our automated interface allows for the evaluation of thousands of samples and multiple model iterations at a negligible cost compared to human annotation. This accessibility encourages broader adoption and more frequent testing, which is critical for advancing the field of text-to-SQL.

In summary, while we trade off a degree of behavioral realism, we gain the reliability, validity, and operability essential for a scientific benchmark. We believe this is the optimal approach for quantitatively assessing underspecification resolution capabilities.

\begin{table*}[bt]\centering
\begin{tabular}{ll}
\hline
Statistics & Num \\ \hline
Non-underspecified & 5,095 \\
SELECT underspecification & 197 \\
FROM underspecification & 3,274 \\
WHERE underspecification & 4,421 \\
Mixed-underspecification & 12,118 \\
Single intent & 1,966 \\
Multi intents & 23,139 \\
Single Table & 3,851 \\
Multi Tables & 21,254 \\
Slot type & 6 \\
Poperty Information   Tables & 102 \\
Real Estate Enterprise Finance   Information Tables & 4 \\
Land Auction Information   Tables & 103 \\
Train & 14,973 \\
Validation & 5,046 \\
Test & 5,086 \\ \hline
\end{tabular}\caption{Dataset Statistics.}\label{Dataset Statistics}
\end{table*}

\subsection{Dataset Statistics}
\label{appendix:Dataset Statistics}

Table \ref{Dataset Statistics} presents the detailed statistics of the ODTQA-FoRe dataset concerning question underspecification, intent type, and table type. The dataset comprises 5,095 non-underspecified questions (20.29\%). It also features questions with underspecifications related to specific SQL clauses: 197 for SELECT (7.85\%), 3,274 for FROM (13.04\%), and 4,421 for WHERE (17.61\%). Notably, the dataset contains 12,118 questions with mixed underspecifications (48.27\%), underscoring the complexity of the challenge. Regarding intent, it includes 1,966 single-intent and 23,139 multi-intent questions. Based on table sources, there are 3,851 single-table and 21,254 multi-table questions, increasing the reasoning complexity. The dataset features 6 distinct slot types and is built upon 102 Property Information Tables, 4 Real Estate Company Finance Information Tables, and 103 Land Auction Information Tables. To facilitate model development and evaluation, the dataset is partitioned into a training set (14,973), a validation set (5,046), and a test set (5,086). These statistics underscore the comprehensive and representative nature of ODUTQA-MDC, especially in terms of the complexity of underspecification and the diversity of annotation labels.

\section{Supplementary experiments}

\subsection{SLU Task}
\label{appendix:SLU Task}

Table \ref{SLU compare} presents the performance comparison of Intent and Slot prediction in queries using BERT and ICL methods. For the BERT model, we fine-tune the ``Bert-base-chinese\footnote{https://huggingface.co/google-bert/bert-base-chinese}'' version on the training set, following the procedure outlined in the main text. The training hyperparameters for BERT are a learning rate of $ 1\times 10^{-5}$ and a batch size of 16.
In the ICL scenario, we select a sufficient number of samples from the training set, covering all intent types, to construct the context. These samples included both intent and slot information, enabling LLMs to generate intent and slot labels for new queries in a single pass. When the SLU label format in the LLM output is not standardized, we select the label that most closely matches the output characters. The experimental results indicate that the fine-tuned BERT model achieves better performance in predicting SLU labels compared with ICL methods. However, fine-tuning requires extensive manual annotation, whereas the ICL methods can achieve comparable results with only a few examples.

\subsection{Implementation of baseline}
\label{appendix:Implementation of baselin}

Furthermore, to highlight the challenges our dataset presents to existing methods, we adopt SLUTQA as a baseline approach. SLUTQA follows the design described in its original paper, selecting prompt templates based on the detected user intent and incorporating SLU information into the prompt to guide the LLM in generating more accurate SQL. Notably, SLUTQA does not invoke the clarification interface, as it lacks a built-in mechanism for underspecification detection.

\subsection{Ablation Study}
\label{appendix:Ablation Study}

To evaluate the contribution of each module within the MAIC-TQA framework in handling underspecified queries, we conduct a series of ablation studies on the Qwen3 32B and Qwen3 30B A3B models. These experiments involve removing the clarification mechanisms for intent underspecification (SELECT), scope underspecification (FROM), and condition underspecification (WHERE). The results are presented in Table \ref{Ablation Study}.

The results indicate that removing the intent clarification mechanism has the least impact on performance. Specifically, for Qwen3 32B, ECR and EA decrease by 6.13\% and 4.45\%, respectively. For Qwen3 30B A3B, ECR and EA decrease by 3.69\% and 4.81\%, respectively. In contrast, removing the scope clarification mechanism most severely affects the ECR metric, leading to a significant drop of 16.92\% for Qwen3 32B and 15.31\% for Qwen3 30B A3B. Meanwhile, removing the condition clarification mechanism has the most pronounced impact on the EA metric, causing it to fall by 28.59\% and 27.29\% for the two models. These findings highlight that scope and condition underspecifications are the primary factors limiting performance on this dataset and confirm the effectiveness of our proposed clarification mechanisms in addressing these issues.

\subsection{Impact of Dynamic Clarification.}

As shown in Tables \ref{tab:underspec_overall}, different models exhibit varying performance fluctuations between fixed and dynamic clarification modes. It is important to clarify that this variance is not due to interface design flaws or the provision of incorrect ("bad") clarification information. Instead, our interface employs a strict built-in verification mechanism to ensure reliability: it retrieves canonical fact information from the ground-truth dictionaries \textit{only} when the system's detection label is correct. In dynamic mode, while the LLM rewrites this canonical information into conversational paraphrases (e.g., using colloquialisms or inverted structures), the design strictly enforces the inclusion of key entity words. This ensures that the generated content remains semantically fact-accurate. Therefore, the observed performance differences are primarily attributed to the varying robustness of different backbone models against diverse linguistic surface forms, as well as the inherent stochasticity in generation, rather than factual noise introduced by the clarification interface. Notably, our main conclusions remain stable across both modes, as MAIC-TQA consistently demonstrates significant superiority.

\begin{table}[]\centering
\begin{tabular}{c|c|ccc}
\hline
Model & Task & P & R & F1 \\ \hline
\multirow{2}{*}{BERT} & Intent & 98.51 & 98.61 & 98.56 \\
 & Slots & 99.68 & 99.88 & 99.78 \\ \hline
\multirow{2}{*}{Qwen3 30B} & Intent & 90.57 & 91.73 & 91.00 \\
 & Slots & 97.01 & 96.79 & 96.90 \\ \hline
\multirow{2}{*}{Qwen3 32B} & Intent & 88.27 & 91.27 & 89.74 \\
 & Slots & 96.07 & 97.23 & 96.64 \\ \hline
\end{tabular}\caption{Performance of Different Models on SLU Tasks.}\label{SLU compare}
\end{table}

\begin{table}[]
\begin{tabular}{cc|cc}
\hline
Method & Model & ECR & EA \\ \hline
 & Qwen3 32B & 97.46 & 57.44 \\
\multirow{-2}{*}{MAIC-TQA} & Qwen3 30B & 96.32 & 54.11 \\ \hline
 & Qwen3 32B & 91.33 & 52.99 \\
\multirow{-2}{*}{w/o SELECT} & Qwen3 30B & 92.63 & 49.30 \\ \hline
 & Qwen3 32B & 80.54 & 35.98 \\
\multirow{-2}{*}{w/o FROM} & Qwen3 30B & 81.01 & 31.84 \\ \hline
 & Qwen3 32B & 90.72 & 28.85 \\
\multirow{-2}{*}{w/o WHERE} & Qwen3 30B & 90.34 & 26.82 \\ \hline
\end{tabular}\caption{Ablation study of the clarification modules in the MAIC-TQA framework.}\label{Ablation Study}
\end{table}

\section{Computing Infrastructure Statement}
\label{appendix:Computing Infrastructure Statement}

All neural network models are implemented using PyTorch\footnote{https://pytorch.org/} v2.3.1. For training the BERT model, we use a single NVIDIA GeForce RTX 4090 GPU.

For experiments of LLMs, we perform inference with the SGLang library\footnote{https://docs.sglang.ai/} on eight NVIDIA A800-SXM4-80GB GPUs. Specifically, we allocate four GPUs to the Qwen3 32B model, four to the Qwen3 30B A3B model. 
Model responses for Kimi K2 (0711-preview version), GLM 4 plus, and Doubao Seed 1.6 (flash version) are obtained through their respective official APIs
\footnote{Kimi K2: \url{https://api.moonshot.cn/v1}}
\footnote{GLM 4 plus: \url{https://open.bigmodel.cn/api/paas/v4}}
\footnote{Doubao Seed 1.6: \url{https://ark.cn-beijing.volces.com/api/v3}} to ensure consistency and reliability of the inference process.

\section{Prompts and Functions of Agent}
\label{appendix:Prompts and Functions of Agent}

\begin{figure*}[htbp]
 \centering
 \includegraphics[width=\textwidth]{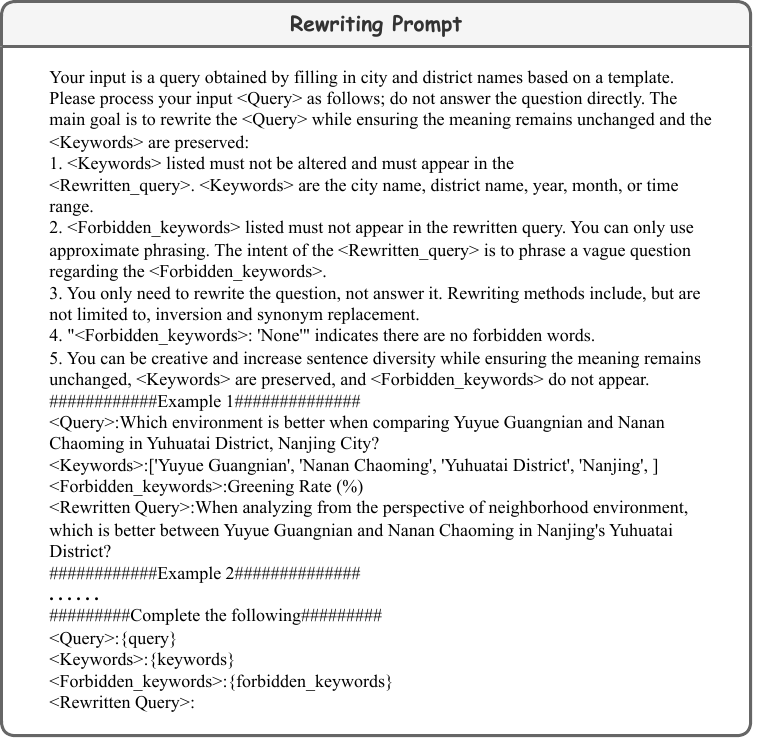}
 \centering
 \caption{Query rewriting prompts for dataset construction.}
 \label{fig:rewrite prompt}
\end{figure*}

This section summarizes the prompts used for data construction and within the MAIC-TQA framework. The rewriting prompt for data construction is presented in Figure \ref{fig:rewrite prompt}.

\begin{figure*}[ht]
 \centering
 \includegraphics[width=\textwidth]{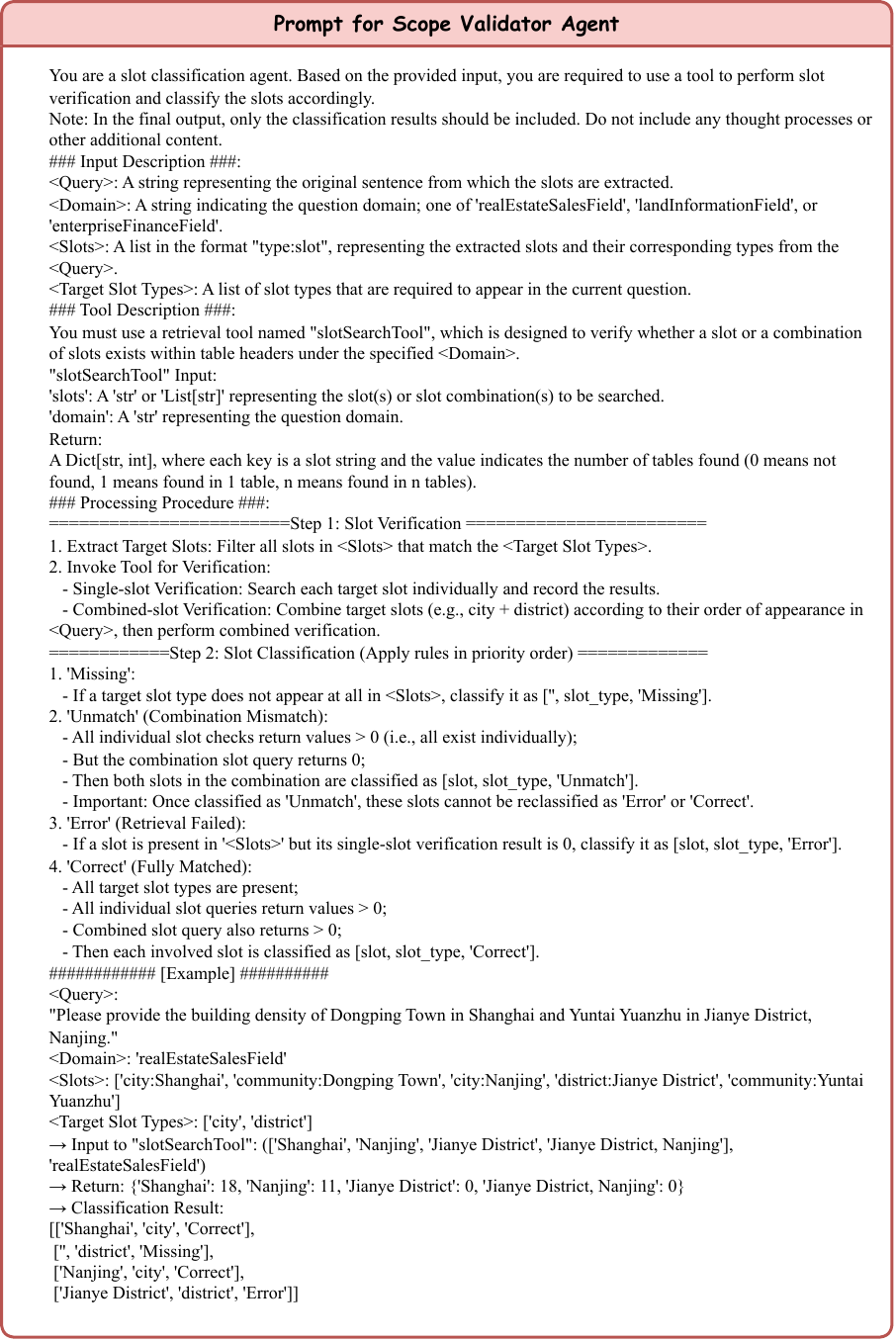}
 \centering
 \caption{Prompt for slots classification.}
 \label{fig:SV_Agent_prompt}
\end{figure*}

\begin{figure*}[htbp]
 \centering
 \includegraphics[width=\textwidth]{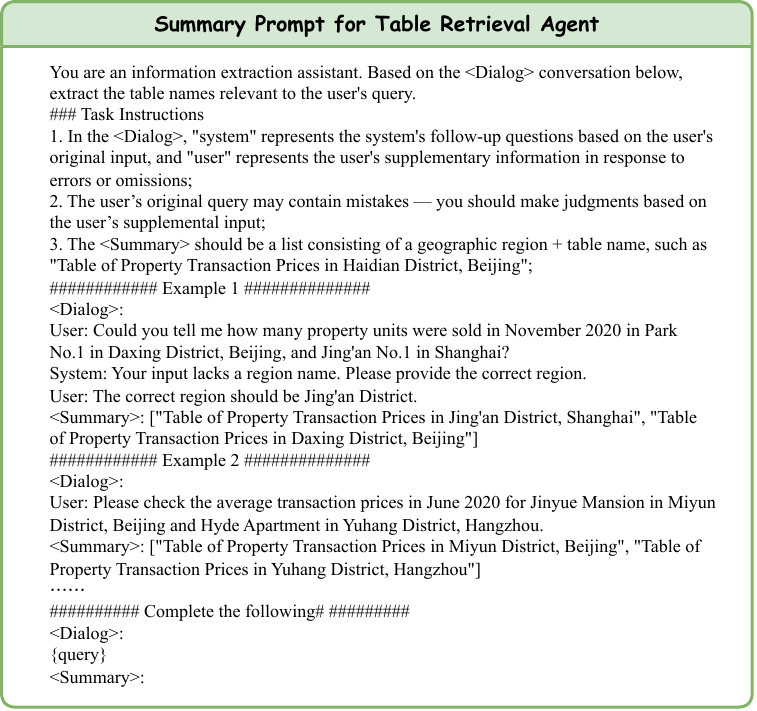}
 \centering
 \caption{Prompt for table caption summarization.}
 \label{fig:summary prompt}
\end{figure*}

\begin{figure*}[ht]
 \centering
 \includegraphics[width=\textwidth]{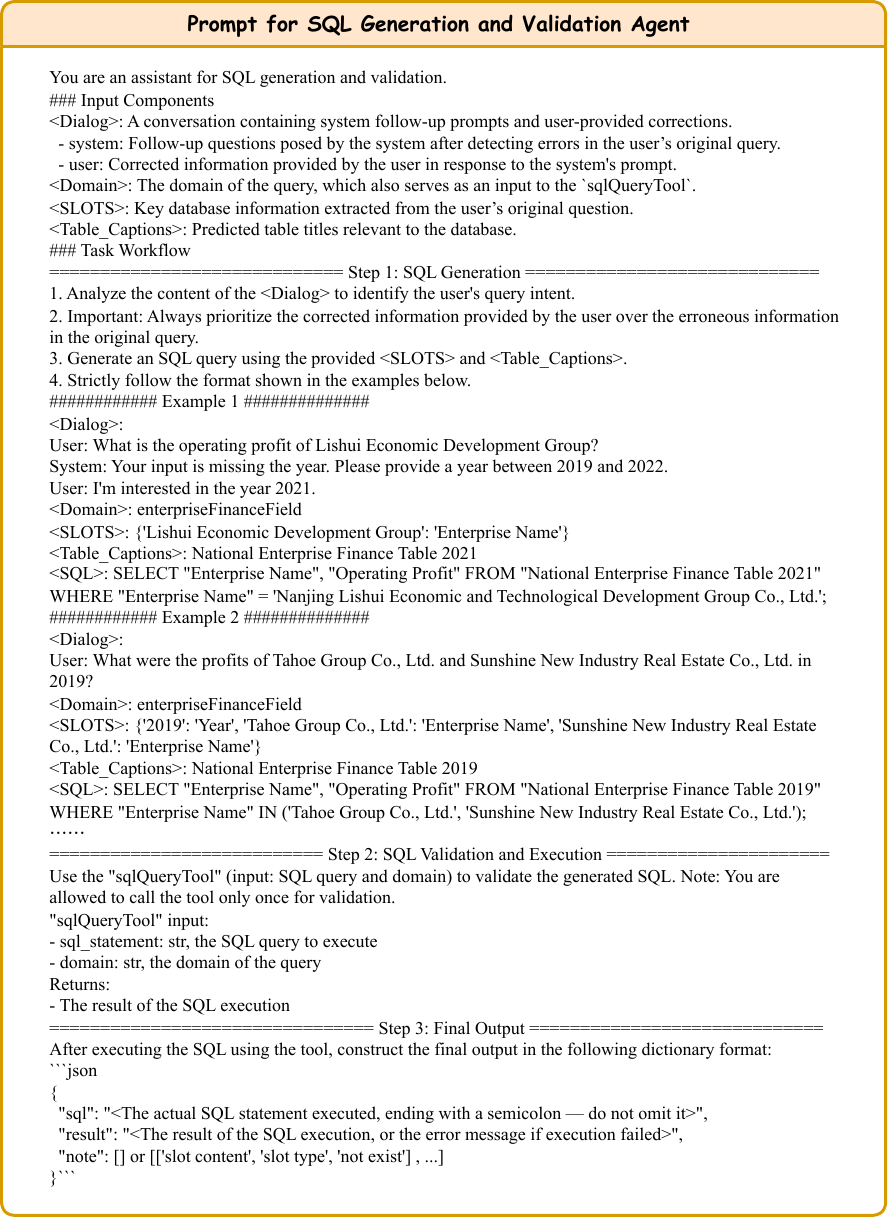}
 \centering
 \caption{Prompt for SQL generation and condition underspecification detection.}
 \label{fig:SQL-GV_Agent_prompt}
\end{figure*}

Within the MAIC-TQA framework, the Scope Validator Agent uses the prompt shown in Figure \ref{fig:SV_Agent_prompt} to verify the scope information in user queries, utilizing the Slot Retrieval Tool defined by Algorithm \ref{alg:slotSearchTool}. The Table Retrieval Agent then uses the prompt shown in Figure \ref{fig:summary prompt} to summarize the target table. Finally, the SQL Generation and Validation Agent utilizes the prompt in Figure \ref{fig:SQL-GV_Agent_prompt} to generate and validate SQL queries, and utilizes the tool defined by Algorithm \ref{alg:sqlQueryTool} to execute SQL statements.

\begin{algorithm*}[h]
\caption{The slotSearchTool used by the SV agent}\label{alg:slotSearchTool}
\begin{algorithmic}[1]

\REQUIRE $slots$: str or List[str]; $domain$: str
\ENSURE $result$: Dict[str, int]
\PARAMETER $all\_domains$: List[str]

\IF{$domain \notin all\_domains$}
    \STATE Raise error: ``Unknown domain"
\ENDIF

\IF{$slots$ is a string}
    \STATE Convert $slots$ to a list with one element
\ENDIF

\STATE $captions \gets$ all\_table\_captions[domain]
\STATE Initialize $result \gets$ empty dictionary

\FOR{each $slot$ in $slots$}
    \STATE $count \gets 0$
    \FOR{each $caption$ in $captions$}
        \IF{$slot$ is a substring of $caption$}
            \STATE $count \gets count + 1$
        \ENDIF
    \ENDFOR
    \STATE $result[slot] \gets count$
\ENDFOR

\RETURN $result$

\end{algorithmic}
\end{algorithm*}

\begin{algorithm*}[h]
\caption{The sqlQueryTool used by the SGV agent}\label{alg:sqlQueryTool}
\begin{algorithmic}[1]

\REQUIRE $sql\_statement$: str; $db\_name$: str
\ENSURE $result$: str
\PARAMETER $all\_db\_names$: List[str]
\STATE Define $try\_execute\_on\_db(database)$:
\STATE \hspace*{1.5em} Create PostgresQueryExecutor instance with given $database$
\STATE \hspace*{1.5em} Return result of executing $sql\_statement$
\IF{$db\_name \in all\_db\_names$}
    \STATE $result \gets try\_execute\_on\_db(db\_name)$
    \STATE \RETURN str($result$)
\ELSIF{$db\_name ==$ `unknowDomain'}
    \FOR{each $db\_name$ in $all\_db\_names$}
        \STATE $result \gets try\_execute\_on\_db(db\_name)$
        \IF{$result$ is of type list}
            \STATE \RETURN str($result$)
        \ENDIF
    \ENDFOR
    \STATE \RETURN ``Failed to execute SQL on all databases."
\ELSE
    \STATE \RETURN ``Invalid database name: " + $db\_name$
\ENDIF
\end{algorithmic}
\end{algorithm*}

\end{document}